\documentclass[letterpaper]{article} % DO NOT CHANGE THIS
\usepackage[]{aaai25}  % DO NOT CHANGE THIS
\usepackage{times}  % DO NOT CHANGE THIS
\usepackage{helvet}  % DO NOT CHANGE THIS
\usepackage{courier}  % DO NOT CHANGE THIS
\usepackage[hyphens]{url}  % DO NOT CHANGE THIS
\usepackage{graphicx} % DO NOT CHANGE THIS
\usepackage{natbib}  % DO NOT CHANGE THIS AND DO NOT ADD ANY OPTIONS TO IT
\usepackage{caption} % DO NOT CHANGE THIS AND DO NOT ADD ANY OPTIONS TO IT
\usepackage{algorithm}
\usepackage{algorithmic}
\usepackage{booktabs}
\usepackage{multirow}
\usepackage{makecell}
\usepackage{xcolor}
\usepackage{array}
\usepackage{newfloat}
\usepackage{listings}
\DeclareCaptionStyle{ruled}{labelfont=normalfont,labelsep=colon,strut=off} % DO NOT CHANGE THIS
\floatstyle{ruled}
\newfloat{listing}{tb}{lst}{}
\floatname{listing}{Listing}
\nocopyright %-- Your paper will not be published if you use this command
\title{Why Do Explanations Fail? A Typology and Discussion on Failures in XAI}
\author{
    Clara Bove\textsuperscript{\rm 1}\equalcontrib,
    Thibault Laugel\textsuperscript{\rm 1, 2}\equalcontrib,
    Marie-Jeanne Lesot\textsuperscript{\rm 2},
    Charles Tijus\textsuperscript{\rm 3}, Marcin Detyniecki\textsuperscript{\rm 1, 2, 4}
}
\affiliations{
    \textsuperscript{\rm 1} AXA, Paris, France \quad
    \textsuperscript{\rm 2} TRAIL, LIP6, Sorbonne Universite, Paris, France
    \\
    \textsuperscript{\rm 3} Laboratoire CHArt-Lutin, Universite Paris 08 Paris, France
    \textsuperscript{\rm 4} Polish Academy of Science, IBS PAN, Warsaw, Poland\\
    contact: thibault.laugel@axa.com
}

\begin{document}

\maketitle

\begin{abstract}
  As Machine Learning models  achieve unprecedented levels of performance, the XAI domain aims at making these models understandable by presenting end-users with intelligible explanations.
    Yet, some existing XAI approaches fail to meet expectations: several issues have been reported in the literature, generally pointing out either 
    technical limitations or misinterpretations by users.
    In this paper, we argue that the resulting harms arise from a complex overlap of multiple failures in XAI, which existing ad-hoc studies fail to capture.
    This work therefore advocates for a holistic perspective, presenting a systematic investigation of limitations of current XAI methods and their impact on the interpretation of explanations. %
    By distinguishing between system-specific and user-specific failures,
    we propose a typological framework that helps revealing the nuanced complexities of explanation failures.
    Leveraging this typology, we discuss some research directions to help practitioners better understand the limitations of XAI systems and enhance the quality of ML explanations.    
\end{abstract}

\section{Introduction}

The field of eXplainable Artificial Intelligence (XAI)  aims at addressing the challenge of providing users with explanations regarding decisions of Machine Learning (ML) models, bridging the gap between the inner workings of complex algorithms and human understanding. 
It constitutes a multidisciplinary domain, drawing upon not only computer science but also cognitive sciences, philosophy and human-computer interaction~\cite{miller2019explanation,byrne2023good,Liao2020,molnar2020interpretable,zednik2021solving}.

Central to the study of XAI is the elusive concept of a "good" explanation. 
Formal definitions, see e.g.~\cite{amgoud2022axiomatic}, do not capture the human component and prove to be a daunting task, leading researchers to gradually develop ad-hoc desiderata and investigate specific challenges that arise. 
The domain multidisciplinarity often results in fragmented investigations, with different research communities focusing on disparate aspects of the problem. On one side, some limitations of explainers such as their lack of robustness~\cite{alvarez2018robustness} or faithfulness~\cite{jacovi2020towards,XAI_nlp_survey} have been mostly investigated by the ML community. In parallel, other works in different fields (e.g. social sciences) have pointed out issues such as the difficulty for explanations to meet user needs~\cite{matarese2021}, prior beliefs~\cite{riveiro2022} or general reasoning~\cite{bertrand2022}. 
 However, this fragmented approach poses a significant challenge, as the limitations that XAI systems face are not mutually exclusive; we argue that they may overlap and conflate with each other.
Consequently, AI practitioners may find themselves unable to comprehend the origin of a failure, and thus to properly mitigate the resulting harms. 

To circumvent this issue, we argue that it is necessary to go beyond existing ad-hoc and domain-specific discussions on XAI issues %,%, which generally focus on either the technical (e.g., \cite{molnar2020general}) or the user perspective (e.g., \cite{bertrand2022}), 
and adopt a holistic approach to XAI failures. Leveraging existing works on XAI approaches, interfaces and evaluations, the main contribution of this paper is the first typology that encompasses insights from both the XAI-ML and the XAI-HCI (Human-Computer Interaction) communities, thereby directly accounting for the complex multidisciplinarity of the XAI field.  
%Going beyond existing discussions on XAI issues that generally focus only on either the technical (e.g.~\cite{molnar2020general}) or the user~\cite{bertrand2022}perspective, 
We provide a systematic and hierarchically organized overview of XAI failures, distinguishing between \emph{system-specific} and \emph{user-specific} ones. Contrary to a systematic literature review, our goal is not to cover all existing works, but rather discuss, for each failure, their origin, characteristics, and some potential mitigation solutions from the literature. 
We believe that this typology can help AI practitioners gain a deeper understanding of their connections and origins. 
Leveraging this typology, we then bring together the system-centric and user-centric perspectives to discuss research avenues to enhance the quality of explanations provided by XAI systems. By fostering a more nuanced understanding of the limitations inherent to XAI, we hope to pave the way for more effective and transparent automated decision-making processes.

This paper is organized as follows: after describing in Section~\ref{sec:background} 
%the necessary background for our work, presenting 
the context and the relation to existing works, we discuss in Section~\ref{sec:rq-and-method} our research objectives, as well as the methodology followed to build the proposed typology of XAI failures.
%context and motivation for a holistic view of XAI failures, and provides an overview of the typology of XAI failures we propose. 
The latter is then presented in two sections corresponding to its root split:  Section~\ref{sec:typology_syst} discusses the system-specific failures and Section~\ref{sec:typology_user} the user-specific ones. Finally, we discuss in Section~\ref{sec:discussion} insights provided by the typology, answering our research questions.

%\section{Background}
\section{Context and Related Works}
\label{sec:background}

%\mj{pas deux titres qui se suivent : il faut mettre un mini chapeau, au moins une phrase qui annonce le contenu de la section}
%\subsection{Setting}
This background section  successively discusses  the  setting we consider  and related works. 
%. First, we frame the setting we consider in Section~\ref{sec:setting}. Then, we discuss related works in Section~\ref{sec:related-works}. 

\subsection{Explanation Process Components}
\label{sec:setting}
%Throughout the paper, we consider that the explanation process is composed of two components of different nature: (i)~the ML system and (ii)~the users. 
The explanation process is generally seen as being composed of two components of different nature: the Machine Learning system on one hand and the user on the other hand.

%\paragraph{The Machine Learning System}
The \textbf{Machine Learning system} (ML system) is in turn composed of two components, that may be entangled and difficult to distinguish: the ML model, that provides decisions, and the explanation method, that  generates rationale for these predictions. We consider a supervised learning context, where given some input information, a ML model returns an associated decision. The data used as input can be either structured (e.g. tabular) or unstructured (images, text, etc.). The nature of the model can vary over a wide range, from simple (e.g. linear) models to deep neural networks or large language models.
The model performance can be evaluated using various metrics of performance, e.g. accuracy or computational complexity to name two examples. 
%; in the paper we focus on accuracy. 
In addition to the prediction itself, the system provides rationale for it, through an explanation generator. It can either be the predictive model itself (in the case a transparent model is used) or a separate system, composed of one (or several) explainer(s),
% called post-hoc, 
built on top of the predictive model. 
There exists a huge diversity of methods to generate various kinds of information that act as explanations (see e.g. \citet{dwivedi2023explainable} for a recent survey), either for one prediction (local explanation) or for the whole model behavior (global explanation). Additionally, several works propose to design eXplanation User Interfaces (XUIs), see e.g. ~\citet{chromik2021human}, to display the generated explanation to the end-users in an intelligible and useful manner.

%\paragraph{The users}
The counterpart within this two-part explanation process is the \textbf{user} who receives the explanation and interacts with the ML system to accomplish their task. The explanation should allow them to understand the decisions made by the model. It must be underlined that users can have various objectives, e.g. depending on their expertise levels and prior knowledge \cite{Liao2020}, to name two examples, that can lead to different needs in terms of interpretability, see e.g.~\cite{mohseni2018}. 

\paragraph{Notion of Explanation Failure} As a consequence of this explanation process structure, it can be argued that 
% In this explanation process, we consider that 
a successful explanation depends on three elements: the ability (i)~of the ML model to make an accurate prediction, (ii) of the explainer to provide a faithful explanation that addresses user needs, and (iii) of the user to properly understand and use it. When at least one of these elements fails, we say that there is an \emph{explanation failure} that needs to be investigated.

\subsection{Related Works}
\label{sec:related-works}

There is an abundant literature focusing on XAI limitations, that we detail in the next sections. However, the vast majority of these works 
%focus on addressing
address such failures by adopting a technical point of view, as opposed to a human-in-the-loop approach, see e.g. \citet{molnar2020general,arrieta2020,srivastava2022xai,saeed2023explainable,bodria2023benchmarking} for some overviews. 
In comparison, very few works suggest that limitations stemming from the user side should also be investigated, such as mismatches between explanations and user needs~\cite{matarese2021} or cognitive biases~\cite{bertrand2022}.

In addition, most existing works consist in ad hoc studies on specific problems, viewing them as independent from one another. This is further exacerbated by the fact that 
technical failures and issues on the user side of the explanation process are generally studied in different domains (Computer Science for the first, Human Computer Interaction and Cognitive Sciences on the other). 
Yet, given the interactive nature of the explanation process~\cite{hilton1990conversational}, it is likely that addressing issues in a more global manner is needed: the process of an explanation is sequential, a "conversation" between the ML system first providing predictions and explanations and then the user interpreting and possibly interacting with them~\cite{miller2017explainable,miller2019explanation}. It can therefore be expected that some failures may interact, conflate, or even amplify one another, raising the need for a holistic perspective on XAI issues. While some contributions in this direction have been proposed, they remain generally focused on domain-specific contexts~\cite{vellido2020importance,antoniadi2021current}.

\section{Research Questions and Methodology}
\label{sec:rq-and-method}

This section describes the methodology we implement to build the proposed typology on XAI  failures, %adopting a global perspective, 
after discussing the research questions we identify. 

% In this paper, we propose to adopt a global perspective on XAI failures, encompassing issues on both the technical and the user side. In this section, we detail how we build this contribution, starting with identifying relevant research questions.

\subsection{Research Questions}
\label{sec:rq}

%We focus in this work on answering the following research questions (RQ):

As stated in the previous sections, the main research question we address can be formulated as follows: 
%\begin{itemize}
    %\item 
    \paragraph{RQ1: What are the different failures that may arise during the explanation process?} 
%\end{itemize} 
A related valuable information concerns the risks the failures can lead to. In this paper, we do not consider the issue of malicious uses of explanations or deliberate intent to fool explanation systems, and the domain of \emph{deceptive XAI}~\cite{dimanov2020you,lakkaraju2020fool,slack2021counterfactual,schneider2023deceptive}, to be a an explanation failure in itself. However, we include a discussion on the explanation dysfunctions that may be exploited by potential malicious actors. 

The identification, and structuration, of failures can help on the way to propose detection and mitigation strategy, which leads to the following second research question. 
%. In order to support such an aim, we consider the following second research question.  
%\begin{itemize}
    %\item 
    \paragraph{RQ2: How can these failures be avoided?}
%\end{itemize} 
To answer this question, it is crucial to understand how failures happen and possibly the reasons why they can occur. The typology we propose thus includes discussions regarding these topics. 

% Past research has exposed numerous failures, but it remains unclear how these are connected, and what their associated risks are.

%\begin{itemize}
%    \item \emph{RQ1: What are the different failures that may arise during the entire explanation process?} Past research has exposed numerous failures, but it remains unclear how these are connected, and what their associated risks are.
%    \item \emph{RQ2: How can these failures be avoided?} It is crucial to understand how these failures happen in order to propose detection and mitigation mechanisms.
%\end{itemize}

% To answer these questions, we propose to build a typology of XAI failures. 

In order to answer these questions, we apply 
%the paper-guided methodology described below, 
a paper-guided approach, following the methodology described in the next subsection, 
basing the proposed typology on the analysis of related publications. However, contrary to a systematic literature review, our goal is not to cover all existing works and to provide an exhaustive survey, but to propose a categorization of failures identified in the literature.

\begin{table*}[]
    \centering
    \resizebox{0.98\linewidth}{!}{
    \begin{tabular}{|c|c|c|c|}
        \hline
        Meta-characteristic & Failure name & \makecell{Discuss the failure or its consequences (\emph{Why it happens?} and \emph{Why is it a problem?)}}  & Discuss solutions \\% & Total distinct \\
        %\midrule
        %\midrule
        \hline \hline
        \multirow{4}{*}{System-specific} & Misleading & \makecell{ \citet{laugel2019issues,ye2022unreliability,papenmeier2019model}\\
        \citet{jacovi2020towards,laugel2018defining,han2023ignorance}\\
        \citet{kaur2020interpreting,agarwal2024faithfulness,colin2022cannot}} & \makecell{\citet{jacovi2020towards,laugel2018defining} \\ \citet{han2023ignorance,agarwal2024faithfulness}\\
        \citet{li2023mathcal}
        }
        \\% & 8 \\
        \cline{2-4}
        %\cmidrule(){2-4}
        & Competing & \makecell{\citet{gosiewska2019not,tsang2020does}\\
        \citet{casalicchio2019visualizing,hooker2021unrestricted}\\ \citet{Suffian2022,bove2022contextualization,laugel2023achieving,zhou2021}\\
        \citet{goethals2023manipulation,zhou2023explain}\\
        \citet{mase2019explaining,mothilal2020explaining}} 
        & \makecell{\citet{aas2021explaining,bove2022contextualization}\\
        \citet{gosiewska2019not,salih2024characterizing}\\
        \citet{jiang2025realexp}
        }\\% & 13 \\
        \cline{2-4}
        & Unstable & \makecell{\citet{jacovi2020towards,alvarez2018robustness,hancox2020robustness}\\
        \citet{slack2020fooling,mishra2021survey,kindermans2019reliability}\\
        \citet{dombrowski2019explanations,ghorbani2019interpretation,zhou2021s}\\
        \citet{sharma2020certifai,molnar2020interpretable,radensky2022exploring}\\
        \citet{visani2022statistical,hickey2021fairness,laugel2019issues}\\
        \citet{zhou2023explain,goethals2023manipulation}} & \makecell{\citet{zhou2021s,zafar2019dlime}\\
        \citet{alvarez2018towards,slack2021reliable}\\
        \citet{dombrowski2019explanations,visani2022statistical}\\
        \citet{gosiewska2019not,yeh2019fidelity}\\
        \citet{shankaranarayana2019alime}} \\%& 19 \\
        \cline{2-4}
        & Incompatible & \makecell{\citet{krishna2022disagreement,okeson2021summarize,bansal2020sam}\\
        \citet{bordt2022post,neely2021order,goethals2023manipulation,kaur2020interpreting}\\
        \citet{swamy2022evaluating,roy2022don,slack2020fooling,reingold2024dissenting}\\
        \citet{aivodji2019fairwashing,laugel2023achieving,bove2022contextualization,garreau2020explaining}\\
        \citet{sundararajan2020many,han2022explanation,poyiadzi2021overlooked}%\\
        }
        & \makecell{\citet{roy2022don,bove2023investigating,krishna2022disagreement}\\
        \citet{pirie2023agree,schwarzschild2023reckoning}\\
        \citet{bhatt2021evaluating,decker2024provably}}\\% & 19 \\
        \hline
        \multirow{3}{*}{User-specific} & Mismatch & \makecell{ \citet{Liao2020,VANDERWAA2021,dwivedi2023explainable}\\
        \citet{doshi2017towards,miller2019explanation,mohseni2018}\\
        \citet{bhattacherjee2001,kaur2020interpreting,graffmalle2017,ijcai2021p609}\\
        \citet{arrieta2020,Wang2019,matarese2021}} & 
        \makecell{\citet{srivastava2023,zarlengatabcbm}\\
        \citet{matarese2021,byrne2023good}\\
        \citet{riveiro2022,pazzani2022expert}}\\% & 15 \\
        \cline{2-4}
        & Counter-intuitive & \makecell{\citet{riveiro2022,sohn2019,kaur2020interpreting,nourani2021anchoring}\\
        \citet{jimenez2020drug,Collaris2018}\\
        \citet{thagard1989,ebermann2023explainable,dochy1995}\\
        \citet{brod2013,nourani2021anchoring,Suffian2022}\\
        \citet{cabitza2024explanations,palaniyappan2022,nickerson1998}} & \makecell{\citet{Jeyasothy2022,rieger2020}\\
        \citet{Wang2019,lim2025diagrammatization}\\
        \citet{conati2021toward,ross2017right}\\
        \citet{ebermann2023explainable,koh2020concept}
        }\\%& 16 \\
        \cline{2-4}
        & Biased Inferences & \makecell{\citet{hoff2015trust,Liao2020,miller2019explanation}\\
        \citet{eiband2019impact,lai2019human,rozenblit2002}\\
        \citet{Chromik2021,kliegr2021review,pratto1991automatic}\\
        \citet{nourani2021anchoring,bertrand2022,mueller2019}\\
        \citet{Wang2019,furnkranz2020cognitive}} & \makecell{\citet{Cheng2019,Wang2019,bove2022contextualization}\\
        \citet{nourani2021anchoring,he2023knowing}\\
        \citet{he2025conversational}} \\
        \hline

        %}
    \end{tabular}
    }
    \caption{List of the references chosen to illustrate the typology, 
    categorized by failure and type of contribution. Some references appear in several cells of the table.}
    \label{tab:typology_references}
\end{table*}

\subsection{Methodology}
\label{sec:method}

We build the proposed typology using the guidelines developed by~\citet{nickerson2013method} for Information Systems taxonomies, made of 5 steps: (1) Define a meta-characteristic, (2) Specify ending conditions, (3) Identify a subset of objects, (4) Identify common characteristics and group objects, (5) Group characteristics into dimensions to refine typology. Steps 3 to 5 are repeated iteratively until the ending conditions specified in step~2 are met. 
%At each iteration, 
Steps 3 and 4 can be done in this order, called \emph{empirical-to-conceptual} process,  or in the reverse one, called \emph{conceptual-to-empirical}, where 4  is rephrased as "Conceptualize characteristics and dimensions of objects" and 3 as "Examine objects for these characteristics and dimensions". 
Following this principle, we alternate inductive categories extraction from papers and deductive categorization of papers.
%\thibault{ajouter termes inductive et deductive}

\paragraph{Meta-characteristic} The meta-characteristic aims at providing a basis for identifying the other dimensions that the typology will rely on. All following characteristics are then intended to be logical consequences of the meta-characteristic, itself deriving from the research questions and the typology's intended use. As the typology we propose to build aims primarily to cover XAI failures by adopting a holistic perspective covering both the ML system and the user, we use a binary meta-characteristic distinguishing  \emph{system-specific} failures, grouping issues associated to technical limitations of the ML system, from \emph{user-specific} ones, which encompass issues caused by the inferences users make about the provided explanations. %, that are specific to each user.

% \paragraph{Characteristics and dimensions.} From the meta-characteristic, other specificities of XAI failures were identified to define failure types. For the system-specific failures, a first characteristic was whether the failure could be observed by itself or if it manifested through a \emph{contradiction} of some sort (e.g. between two parts of the explanations). The contradiction-based failures were then distinguished depending on  

\paragraph{Ending conditions} We use both objective and subjective criteria proposed by~\citet{nickerson2013method}: the process stops when no new dimension or characteristic has been added in the last iteration. In addition, it stops when the typology is assessed to be concise (at most 10 types of failures), robust (at least 5 papers in each category), comprehensive % {\color{red}{(????)}} %, extendible
and explanatory (the categories 
are 
%should be
easily distinguishable based on the characteristics). 

\paragraph{Data collection} A crucial step 
%for the construction of the typology 
is the collection of works relevant to the topic of XAI failures. This task has been performed in an iterative manner, enriching the set of collected papers through enriched list of search keywords. 
% At each iteration,  3 authors looked independently for works that are relevant to the topic of XAI failures.
Included in the screening scope are the proceedings of the main venues from the fields of AI (ICML, NeurIPS, IJCAI, etc.), HCI (CHI, IUI, etc.), and % conferences specialized on the topic of 
explainability specialized conferences (FAccT, AIES, XAI conference, etc.). Were also considered papers available on ArXiv to scan for potentially unpublished but meaningful contributions.

The initial list of search keywords included terms as \emph{explainability/explanations/explaining, interpretability/interpreting, transparency}, with and without associations with notions such as \emph{failures, problems, risks, pitfalls, inconsistencies}, etc.  %In HCI venues, terms such as \emph{biases}
Iteratively, after identifying new categories in the taxonomy, it was enriched through 
%Once a first set of categories for the taxonomy was proposed, the next  iterations focused on enriching them, proposing 
category-specific keywords such as \emph{stable/unstable/robust explanations}, etc.
%\emph{stable/instable explanations}, \emph{robust explanations}, etc.

In all iterations, the inclusion criterion of the retrieved papers  in the collection imposes that their contributions  
%inclusion criteria for papers was that the contributions 
: (i) identify and discuss pitfalls of existing explainability methods, either from a theoretical or an empirical perspective; (ii) or propose new explanation methods to mitigate specific issues, with quantitative assessments of these results. 

After summarizing the contributions of each paper and documenting the rationale behind their relevance for the typology, the authors discussed together their inclusion. 
%which papers to include.

\subsection{Overview of the Result}

The methodology described in the previous section lead to select a total number of 
108 papers to build the typology. After the typology was built, we check that each type, and in particular each leaf type, is associated with at least 5 papers, so as to ensure it is representative and significant.%, i.e. it does not correspond to a very specific and perhaps anecdotic case.

The selection of considered characteristics and dimensions is derived from the considered meta-characteristic and driven by the considered research questions, related to the aim of avoiding these failures. For system-specific failures, discussed 
%in detail
in Sect.~\ref{sec:typology_syst}, a temporal dimension related to the explanation process is taken into account, to define subtypes depending on the system development step at which dysfunctional behaviors may occur: the ML model itself, the explanation generator or the generated explanation that may contain conflicting pieces of information. The latter is further decomposed, at a third level of the typology,  depending on the source of the conflict.
For user-specific failures, discussed in Sect.~\ref{sec:typology_user}, the structuring dimensions we propose distinguish whether the explanation is rejected or accepted by the user and additionally examine the rejection cause, depending on whether it related to the explanation form or  content. In case of acceptation, a failure can occur in cases where the explanation is actually misunderstood or misused.

%\paragraph{Characteristics and dimensions} ~\mj{à fusionner avec overview et donner plein d'info : typo elle-même = réponse à RQ1, discussions associées = réponses à RQ2 + expliquer la logique de la typo proposée, quelles dimensions la structurent..}
%Deriving from the meta characteristic, various types of failures were quickly identified, their differences relying around their consequence on the user. For system-specific failures, a distinction was thus made between issues that stemmed from the system itself not functioning as expected (\emph{misleading} explanations), and problems originating from confusion created by functioning as intended, but not behavior that lead to misunderstanding on the side of the user, generally due to some pieces of information contradicting (\emph{competing}, \emph{unstable} and \emph{incompatible} explanations).
%For user-specific failures, similarly a distinction was made between issues created by a reject of the explanation due to its form (\emph{mismatch}) or the information it carries (\emph{counter-intuitive}) and explanations accepted but misunderstood or misused. 

In addition, in order to answer more accurately the considered research questions, we propose to enrich each explanation failure type with a discussion along three axes: (i)~why does the failure happen, (ii)~why is it a problem and (iii)~what kind of solutions, if any, have been proposed on the literature, either to measure or inform about the issue, mitigate its negative consequences or even solve it. Regarding~(ii), it can indeed be observed that, depending on the context, a phenomenon can be seen as an issue or not. This can e.g. be related to the fact that even explanation manipulation can be seen as desirable in specific cases:  \citet{slack2020fooling} argue that it can be used as as a method to preserve intellectual property about the classifier,  avoiding to disclose its underlying principle. An overview of the typology, with the references considered to support it, is shown in Table~\ref{tab:typology_references}.

%Building on the types of contributions identified in the Data Collection process, we structured each category of failure with 3 subquestions:
%    \begin{itemize}
%        \item Why does it happen? aeaz azejohazeoh 
%        \item Why is it a problem? If obvious for some failures, it may not be the case for others. More interestingly, depending on the context, similar phenomenon may be seen as issues by some authors, but not by some others. (ex: incomaptible)
 %       \item What are the solutions? For each failure, several works have been generally proposed, either to measure and inform the issue, mitigate its negative consequences, or try to solve it completely. 
%    \end{itemize}

%\paragraph{Overview of the Result}
%Applying the previous methodology leads us to identify 7 categories, structured as 4 system-specific ones and 3 user-specific ones. They are described and discussed in details in the next section and graphically represented on Figures~\ref{fig:system-specific} and~\ref{fig:user-specific}. 

%\section{Results: A Typology of XAI Failures}
%\label{sec:typology}

\section{Proposed Typology: System-specific\\ XAI Failures}
\label{sec:typology_syst}

\begin{figure*}
    \centering
    \includegraphics[width=0.6\linewidth]{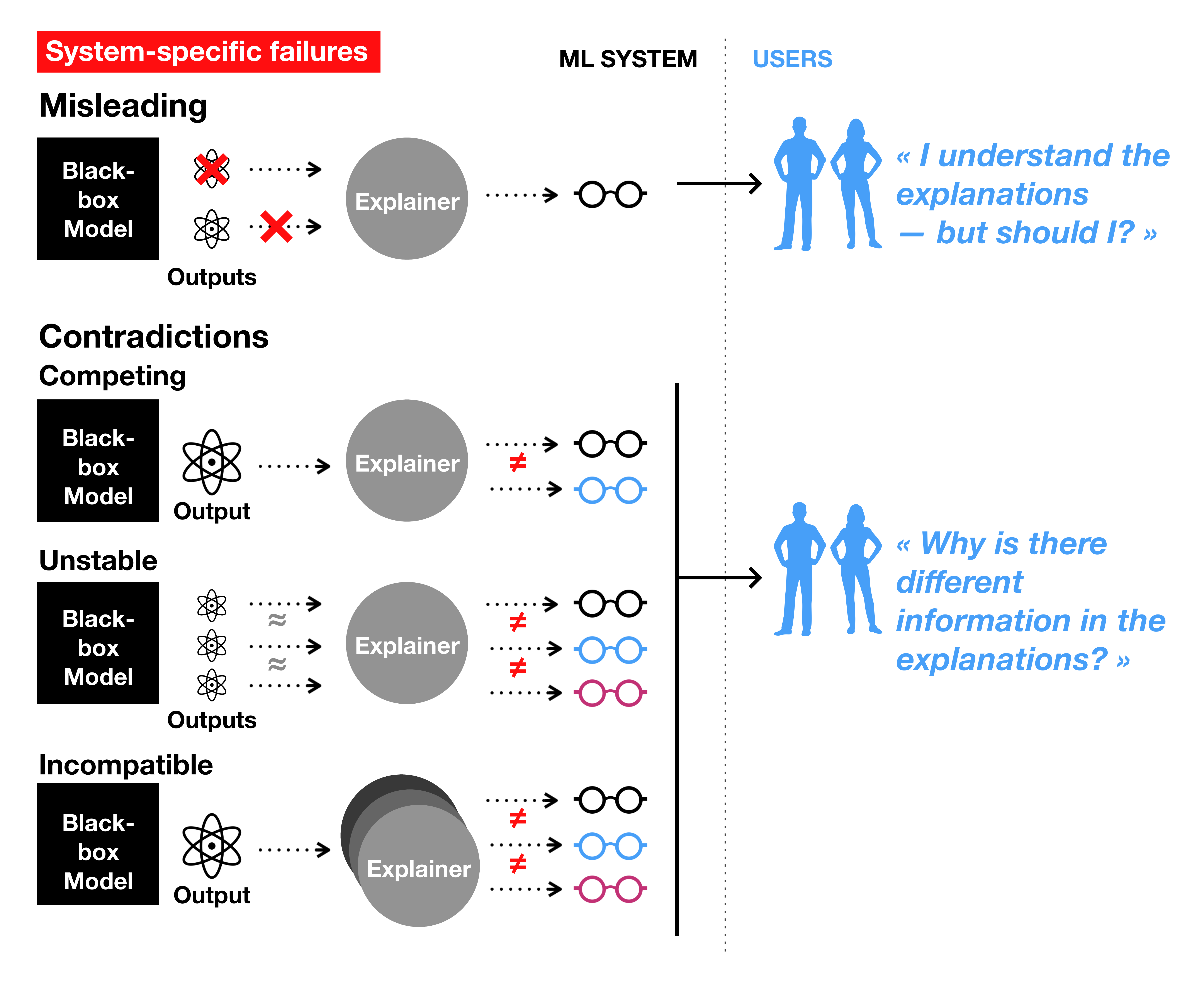}
    \caption{Types of system-specific explanation failures: misleading explanations ascribed to the ML system and  contradicting ones ascribed to the explainer. The cross and the $\neq$ symbol indicate the step at which the failure takes place. }
    \label{fig:system-specific}
\end{figure*}

This section discusses explanation failures that can be ascribed to the Machine Learning system, depending on its development step at which they can occur. A graphical representation of the 4 proposed subtypes, organized in two categories, is provided in 
%We identify system-specific explanation failures as a design problem of the ML system, that may be ascribed to any of its components. The typology built when applying the methodology presented in the previous section is represented 
Fig.~\ref{fig:system-specific} and commented below.
%in the next subsections. 

\subsection{Overview}

The typology  decomposes system-specific explanation failures into two categories:  (1) misleading explanations when either the ML model provides an inaccurate prediction or when the explainer is not faithful; and (2) contradictions when conflicting information are provided by one or several explainers. The effect of the former on the users can be characterized by the summarizing question "I understand the explanations, but should I?". 

The latter can be further decomposed into 3 categories depending on the source of the conflict: inconsistencies can occur because of contradiction between  (a) different pieces of information of the same explanation, leading to explanations we propose to name \emph{competing}, (b) different explanations generated by the same explainer, named \emph{unstable} explanations, or (c) different explanations generated by different explainers, named \emph{incompatible} explanations. This case may occur when the global explainer is defined as a set of explainers. 
In other words, as the diagram in Fig.~\ref{fig:system-specific} illustrates, the plurality that leads to the contradiction can be due to the output, the input or the explainer itself. In all three cases, users may not understand this conflicting information, and wonder "Why is it different?"

\subsection{Misleading Explanations}
\label{sec:typology-misleading}
We call an explanation \emph{misleading} when the failure results from the ML system being dysfunctional, i.e. when it fails to meet the very purpose it was designed for. We distinguish two situations, depending on whether the dysfunction comes from the prediction or the explanation. There is a risk that the explanation, however, could be accepted by the users without their being able to perceive this failure. 

\paragraph{Why does it happen?}
First, it can occur that ML models output confident yet incorrect predictions. %In this case we say the
Such ML models can be said dysfunctional, raising the question of the relevance or potential misleadingness of generating explanations. Second, the XAI system is deemed dysfunctional when it fails to meet the mathematical objectives it has been designed to satisfy. This is related to the notion of unfaithful explanations i.e. that 
%are globally defined as explanations that
fail to  adequately account for the behavior the ML model they are associated to, see e.g. ~\cite{jacovi2020towards,li2023mathcal}. 

Many explanations are not generated through the minimization of a cost function, but are instead defined as closed-form formulas. Examples include saliency maps in Computer Vision~\cite{selvaraju2017grad}, influence functions~\cite{koh2017understanding}, partial dependence plots~\cite{friedman2001greedy,goldstein2015peeking}, or formal explanations~\cite{darwiche20reasons,audemard20tractable,marques2024logic}.
%Most explanations are generated to satisfy required properties pre-defined by the ML practitioner (e.g., optimize certain objectives), either explicitly or implicitly. Some of them, such as gradient-based methods (e.g.,~\cite{selvaraju2017grad}), % or partial dependence plots~\cite{goldstein2015peeking},or formal explanations (e.g.,\cite{darwiche20reasons,audemard20tractable,marques2024logic})  are directly defined as closed-form formulas. 
Such explanations can be considered as 
%In this case, one can consider that they are 
functional by design, and faithful.
%and that the resulting explanations are faithful. %Hence, the issues these methods may encounter fall into other types of failures, detailed later. 

On the other hand, some explanation methods rely on optimizing cost functions, see e.g., counterfactual example generation~\citet{wachter31brent,mothilal2020explaining,laugel2018comparison} or surrogate-based methods~\cite{ribeiro2016should}, and as such do not guarantee that the associated desiderata is satisfied.
% other types of approaches rely on estimating some desired objective, and as such do not guarantee that the associated desiderata is satisfied.\footnote{For instance: (i) counterfactual explainers optimize the distance between the observation to explain and the counterfactual example~\cite{wachter31brent}; (ii) surrogate explainers~\cite{ribeiro2016should} %,baehrens2010explain,craven1995extracting
%may be subject to approximation errors when trying to mimic the behavior of the black-box model.}.
Often, there is no guarantee that a satisfying solution to these problems exists, as discussed for instance by~\citet{ye2022unreliability} for prompt-based explanations for Large Language Models. As a result, the explanation may be unfaithful to the model it aims at explaining.

\paragraph{Why is it a problem?}
When an incorrect prediction is returned, although explanations can be useful for model calibration~\cite{ye2022unreliability}, they may be seen as potentially harmful, especially for users with low levels of awareness~\cite{papenmeier2019model}.
An unfaithful explanation is also obviously problematic, as a lack of fidelity to the model may induce the user either to reject the system (\emph{undertrust}) or to place unwarranted trust in it (\emph{overtrust})~\cite{papenmeier2019model,colin2022cannot}. %,lipton2018mythos

 \paragraph{Solutions.}
 Despite its critical importance, 
 %underlined by numerous discussions in the literature, 
 this faithfulness is still often overlooked in practice, by both users and ML practitioners~\cite{kaur2020interpreting}. 
Even when there is a will to control for faithfulness, the precise definition and evaluation of this notion remains elusive~\cite{jacovi2020towards,laugel2018defining,li2023mathcal} and often at odds with other desired criteria~\cite{han2023ignorance,agarwal2024faithfulness} 
amplifying the challenge.
% Nevertheless, 
Like previous works, we argue that assessing faithfulness is crucial, and that this evaluation should precede all other assessments of the explanation: firstly, contrary to other assessments, it is a purely technical task, allowing ML developers to conduct it independently of end-users. Secondly, it serves as a foundational step for identifying and addressing any other potential issue. In the rest of the paper, the ML system is therefore assumed to be functional, i.e. to provide accurate predictions and faithful explanations.
% Failure to do so poses the threat of conflating faithfulness concerns with other concepts, as illustrated with Example 1 in Sec.~\ref{sec:context}. 

\subsection{Competing Explanations}
\label{sec:typology-competing}
We propose to name  explanations  \emph{competing} when several parts of the explanation are contradicting with one another, if for instance they consist of several counterfactual examples.

\paragraph{Why does it happen?}
 %, as represented in Fig.~\ref{fig:system-specific}. 
We identify two scenarios when this can happen.
Often, explanations are composed of several components, interacting with each other in various ways: e.g., a feature attribution vector represents the contributions of each feature, word or pixel to the prediction.
Yet, numerous works show that more complex effects such as interactions or correlations between features are often not taken into account
% in the calculation of feature importances%\footnote{\cite{jacovi2020towards} refers this as the \emph{Linearity Assumption} of feature attribution methods.}
~\cite{gosiewska2019not,tsang2020does,mase2019explaining,casalicchio2019visualizing,hooker2021unrestricted}.
A contradiction may thus appear between the semantic relationship of two notions, %expected by the user, 
their actual correlation in the data used to train the model, and the explanation returned by the system.
The second scenario is when XAI systems  generate, rather than a single explanation, a set of explanations, e.g. of counterfactual examples~\cite{mothilal2020explaining}
 to provide richer insights to the user. 
In the counterfactual case, we connect  the underlying notion of explanation diversity (see e.g. %the discussion in~
\citet{laugel2023achieving}) to the risk of getting competing explanations: these diverse explanations generally aim at suggesting the user various alternatives, i.e. a choice between several possible actions, but they may appear contradictory. 

\paragraph{Why is it a problem?}
% Whether coming from different explanations among a diverse set, or from diverse individual components within a given explanation, a c
Contradictions between two pieces of information may be perceived as confusing by the user, potentially leading them to  reject the explanation~\cite{Suffian2022}. 
%In the context of an automated insurance pricing scenario, 
For instance \citet{bove2022contextualization} suggest  competing explanations as one of the reasons for which users misunderstand the explanations returned by SHAP.
%For instance  competing explanations have been suggested as one of the reasons for which users misunderstand the SHAP explanations~\cite{bove2022contextualization}.
Furthermore, in a non-cooperative setting~\cite{bordt2022post} where the objectives of the user and the ML developer are not aligned,  this problem can also open up the risk of explanation manipulation through the selection of an explanation that is not in the best interest of the user, see e.g.~\citet{goethals2023manipulation,zhou2023explain} .

\paragraph{Solutions.}
The proposed solutions can be grouped in two categories: the first one focuses on better XAI systems, either by adapting them to take into account correlations~\cite{aas2021explaining,salih2024characterizing}, or enriching them, e.g. by computing,
%proposing to calculate, 
in addition to the usual feature importance vectors, % the contributions of 
feature interactions~\cite{gosiewska2019not,jiang2025realexp}.
A second type of enrichment consists in exploiting expert knowledge to contextualize feature contributions: informing the user, usually through the XUI, may allow to rationalize some confusion that can be caused by competing explanations \cite{bove2022contextualization}. 
%For instance, \cite{bove2022contextualization} propose to use expert knowledge to contextualize feature contributions, thus allowing to rationalize some confusion that may be caused by competing explanations.

%\subsubsection{Unstable explanations}
\subsection{Unstable Explanations}
\label{sec:typology-unstable}
We call explanations \emph{unstable} when there is an inconsistency, i.e. a contradiction, between explanations within a supposedly stable scenario: for instance, when producing local explanations for similar instances with similar outcomes, one might expect that the explanations should be similar as well~\cite{jacovi2020towards}.
Such explanation inconsistencies have been widely observed
%, in particular for post-hoc XAI methods
~\cite{alvarez2018robustness,laugel2019issues,yeh2019fidelity}. %laugel2019issues
Similarly to competing explanations \emph{unstable} explanations are perceived originally as a technical failure of the explainer. Yet, as we describe below, it can also originate from the user, or from a combination of both the system and the user.

\paragraph{Why does it happen?} 
Unstable explanations are often considered as a technical failure of the explainer, viewed as a lack of robustness that needs to be  fixed~\cite{alvarez2018towards,slack2021reliable,mishra2021survey,kindermans2019reliability}. 
However, these inconsistencies can also be ascribed to the model to be explained \cite{dombrowski2019explanations,ghorbani2019interpretation,alvarez2018towards}: %
% MJL : + Guyomard . T: ref?
the local behavior of the latter may indeed vary abruptly, due to the complexity of the task being modeled, the complexity of the model itself or its lack of robustness. Faithful explanations then reflect these steep changes, leading to an apparent lack of stability. 
Moreover, this issue of instability may conflate with user perception of the similarity between instances and the explanations they expect as a result. This similarity, that depends on the user knowledge and possible biases, may differ from the similarity considered by the ML system.
Pushing the expectation of explanation stability to its extreme, explanations generated for identical observations are anticipated to be identical. However,  post-hoc model-agnostic methods, be they local or global (e.g. ~\citet{ribeiro2016should,lundberg2017unified,wachter31brent,altmann2010permutation}),
often rely on a stochastic data generation step~\cite{zhou2021s,visani2022statistical} 
that may cause instability.
%lee2019developing %zafar2019dlime, - T: ????
This comes in addition to the Roshomon effect,  i.e. that several equally good but potentially drastically different solutions can coexist and therefore be selected as explanations~\cite{hancox2020robustness}. 

\paragraph{Why is it a problem?}
Considering that faithful explanations reflect the state of ML model, depending on its potential causes discussed above, a lack of stability can either be seen as an actual failure or as a desired characteristic: for instance more local explanations are expected to be less stable~\cite{molnar2020interpretable,yeh2019fidelity}, with locality being a commonly expressed desideratum for explanations~\cite{radensky2022exploring}. Thus, as for competing explanations, interpreting the lack of stability as a failure depends on the users needs, their knowledge of the explainer and their perception of how similar the explanations should be: a user not being aware of the locality-stability trade-off may see it as problematic but may not otherwise~\cite{hancox2020robustness}.
Still, instability may result in the user rejecting the explanation, or even the whole AI system, seeing it as a proof of unfairness by the model~\cite{sharma2020certifai,hickey2021fairness}, %,
that may be abused by the organization providing the explanation~\cite{goethals2023manipulation,zhou2023explain}.

\paragraph{Solutions.} 
Various approaches have been proposed to address instability, depending on its source.
To fix the stochastic instability of model-agnostic post-hoc explainers, most contributions focus on algorithmic modifications of the random data generation step they rely on, e.g. replacing it with a deterministic one~\cite{zhou2021s,zafar2019dlime}, or through a reweighting strategy~\cite{shankaranarayana2019alime,yeh2019fidelity}.
When instability originates from a misalignment between the user perception and the ML system representation, 
several works propose strategies to constrain the ML model during its training phase~\cite{alvarez2018towards,dombrowski2019explanations}. On a different note, rather than mitigating the problem, several works propose to measure the explanation stability, arguing that it describes a notion of uncertainty that can be helpful for the user to better understand and use them~\cite{gosiewska2019not,shankaranarayana2019alime,slack2021reliable,visani2022statistical}.%

%\subsubsection{Incompatible explanations: the "Disagreement Problem" in XAI}
\subsection{Incompatible Explanations: the "Disagreement Problem" in XAI}
%(or the "Disagreement Problem" in XAI)}
A third contradiction case
%where several explanations can contradict each other 
occurs when several explainers are used in the same setting. We call 
them %such contradictions
\emph{incompatible explanations}, they correspond to the prevalent~\cite{krishna2022disagreement} and well known Disagreement Problem, see e.g.~\citet{sundararajan2020many,han2022explanation,bordt2022post,neely2021order}. %\cite{okeson2021summarize, poyiadzi2021overlooked}

\paragraph{Why does it happen?} 
Assuming that the explanations are faithful and stable, the most high-level root cause of this issue comes from the fact that the task of providing explanations, in particular in the post-hoc setting, is essentially underdetermined~\cite{bordt2022post}: as mentioned in the introduction, the concept of "good" explanation is elusive and has been formalized in numerous different ways, taking into account different types of desiderata. For instance, 
because they rely on different assumptions, 
%the explanations generated by LIME and SHAP 
LIME and SHAP explanations 
are expected to differ, even if both are faithful~\cite{poyiadzi2021overlooked,han2022explanation}. %lundberg2017unified,. 
Similar discussions apply to global feature attribution methods \cite{okeson2021summarize} or counterfactual explanations \cite{goethals2023manipulation}.
Going further, some differences between explanations can  be attributed to discrepancies in the implementations of the supposedly same mathematical explanation objective, see e.g.~\citet{sundararajan2020many} for the case of Shapley value-based explanations. 
Finally, a source of incompatibility can be attributed for the explanations generated using the same method, but different parameters. Indeed, XAI methods generally rely on  hyperparameters that may not always be understood (if known at all) by the user, albeit heavily impacting the obtained explanations~\cite{garreau2020explaining,bansal2020sam}. 
As a result, explanations can differ on multiple bases, for instance, in the case of feature score explanation, ranging from the top features being different to differences in order of importance or direction~\cite{krishna2022disagreement}. 
%On a different note, when one or several of the considered explanations is not faithful or unstable, it is impossible to attribute a potential disagreement as a consequence of the explainers heuristics.

\paragraph{Why (and when) is it a problem?}
This disagreement between explanations is generally viewed as a problem~\cite{sundararajan2020many,garreau2020explaining,poyiadzi2021overlooked,swamy2022evaluating,han2022explanation,bordt2022post,roy2022don,goethals2023manipulation}.
Several user studies have noted it to be a source of confusion for users~\cite{okeson2021summarize,krishna2022disagreement}, resulting in their lowering  trust in the system and therefore possibly leading to a reject of the system as a whole or a questionable selection of the proposed explanations, e.g. based on method popularity~\cite{kaur2020interpreting}.
%In other cases, supposing that all the considered explanations are faithful and are all understood, users have been shown to select an explanation based on some arguably arbitrary criteria such as their own heuristic, or the popularity of the XAI method~\cite{kaur2020interpreting,krishna2022disagreement}.
As for competing explanations, in a non-cooperative setting, this incompatibility between explanations may 
%also be seen as a degree of freedom that a malicious AI practitioner may leverage 
be leveraged by a malicious AI practitioner  to rationalize unfair decisions by choosing the explanation most aligned to their objectives~
\cite{slack2020fooling,aivodji2019fairwashing,goethals2023manipulation}. %aivodji2019fairwashing,goethals2023manipulation 

On the other hand, similarly to the unstable and competing cases, incompatible explanations, if faithful, can be viewed as an opportunity that  may be leveraged for a better interaction with the system in a collaborative context. Indeed, explanation disagreement can be seen as a source of diversity, as noted by~\citet{laugel2023achieving,goethals2023manipulation}, which is viewed positively and helps understanding, as empirically shown by~\citet{bove2022contextualization} when combining counterfactual with global feature attributions. Other works have also leveraged explanation disagreement to reduce user overreliance to the model~\cite{reingold2024dissenting}.
In other cases, disagreements between feature importance explanations are seen as an evidence of a lack of robustness on the side of the classifier, therefore used as a starting point for model auditing~\cite{okeson2021summarize} to improve its performance. 
%Approaches generating plural explanations generally aim for diversity~\cite{mothilal2020explaining,laugel2023achieving} to provide the user several alternatives among which they may intentionally choose their preferred solution, as it has been shown to enhance subjective and objective user understanding~\cite{bove2023investigating}.  

\paragraph{Solutions.}
Intuitively, circumventing the issue of incompatibility may be simply done by hiding disagreements~\cite{roy2022don}. On the contrary, other works propose to emphasize them through interfaces~\cite{bove2023investigating}, sometimes arguing, like for stability, that the level of disagreement may be used as a measure of uncertainty to validate parts of the explanations~\cite{krishna2022disagreement}. 
Using the same idea, other works propose to aggregate various explanation methods~\cite{bhatt2021evaluating,pirie2023agree,decker2024provably} %pirie2023agree
to provide more robust explanations, or even propose to train new models that minimize this disagreement~\cite{schwarzschild2023reckoning}.

%\subsection{User-specific failures}
%\label{sec:typology-userbased}
\section{Proposed Typology: User-specific\\ XAI Failures}
\label{sec:typology_user}
\begin{figure*}
    \centering
    \includegraphics[width=0.6\linewidth]{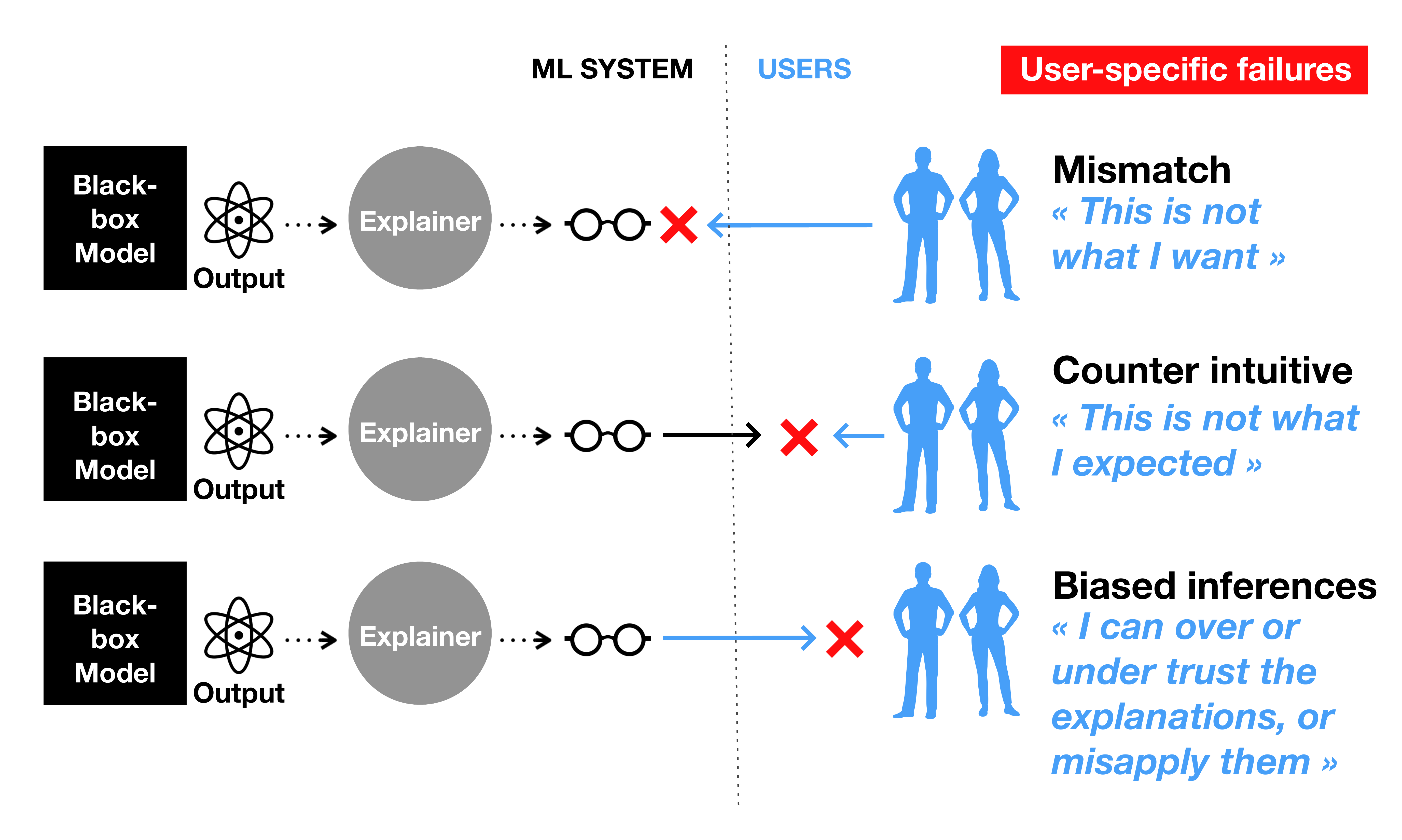}
    \caption{Types of user-specific explanation failures in ML explanations. The ML model is supposed to give an accurate prediction and the explainer to generate faithful explanations. 
    The cross indicates the step at which the failure takes place.}
    \label{fig:user-specific}
\end{figure*}

This section discusses explanation failures that can occur when users misinterpret the explanations provided by a ML system with no technical failures: 
regardless of their quality, these explanations have been shown to sometimes fail in their explanatory objective \cite{Cheng2019,Wang2019}. 
%It may then be dangerous to deploy AI system explanations in high-stakes settings (e.g., health or justice),  without ensuring that they align with the users. 
We propose here a typology of failures that originates from inconsistent users' inferences, distinguishing between three categories graphically represented in Fig.~\ref{fig:user-specific} and discussed in turn below:  \emph{mismatch} failures when there is a contradiction between the ML explanations and the users' expectations in term of format; \emph{counterintuitive} failures when this contradiction concerns the explanation content; and \emph{biased inferences} failures when cognitive biases inherent to each user interfere with the explanations. 
We believe that these user-specific failures should be known so XAI designers can understand the users’ mental model processes and support their interpretation of the provided explanations.

%\subsubsection{Mismatch}
\subsection{Mismatch}
\label{sec:typology-mismatch}
First, we propose to define \emph{mismatched explanations} when the format of the extracted information does not meet the users’ expectations towards the ML system, leading to the remark "This is not what I want" in Fig.~\ref{fig:user-specific}.
%These expectations can vary depending on the context of the interaction, the background knowledge of the users or the output of the model, which makes it challenging to select the adequate XAI methods.

\paragraph{Why does it happen?}
Depending on the context of the interaction and the nature of the decision model's outputs, users have been shown to have different needs and questions regarding the ML system,  
%and the user questions
that may also vary depending on the model’s output ~\cite{Liao2020,VANDERWAA2021}. 
%and the user questions may also vary depending on the model’s output \cite{Liao2020,VANDERWAA2021}.
%Hence, the explanations generated may vary from one approach to another, and not all of them aim to answer the same user question. The latter usually focus on specific needs, and only a limited number of XAI approaches can be used to answer them. Moreover, most XAI techniques are designed to be ad-hoc, meaning that the explanations techniques does not necessarely take into account the context for which the explanations is seeked by a user \cite{matarese2021}.
In parallel, %as mentioned in the introduction, 
there is a huge diversity in the forms explanations can take (see e.g. \citet{dwivedi2023explainable}), but explanation techniques do not necessarily take into account the context in which the explanations is sought by a user~\cite{matarese2021}. 
%As a consequence, 
Therefore, mismatches between the user questions and the generated explanations can occur, on the explanation type (e.g., feature importance or rules), locality (e.g., local or global), goal (e.g., factual or causal) and on the information complexity (e.g., expressed for ML practitioners or lay users), to name a few.

Such mismatches reflect the lack of user-centricity in the conception of the XAI system. Actually, most XAI approaches are designed without evaluating whether the explanations satisfy the needs of real users \cite{doshi2017towards,ijcai2021p609}. 
% miller2019explanation
Instead, other criteria are used to evaluate the relevance of an XAI approach, such as the visual aspect of the explanations, its popularity or the ease of implementation~\cite{mohseni2018,arrieta2020}. 
Such evaluations can bring to light that they fail to meet these needs: 
for example, during the co-design workshop for an AI-based diagnosis tool~\cite{Wang2019}, many interviewed doctors reported that they would prefer alternative hypotheses (e.g., counterfactual examples) rather than factual explanations (e.g., local feature importance scores) that were initially implemented in the system. 

\paragraph{Why is it a problem?}
Mismatched explanations can lead to dissatisfaction and rejection of the whole ML system: 
it has been demonstrated that differences between initial expectations and actual experiences can affect both the user satisfaction and acceptance of a system \cite{bhattacherjee2001}.
%the Expectation Confirmation Model \cite{bhattacherjee2001} relates the user satisfaction and acceptance of a system with the difference between the users initial expectations and their actual experience. 
Along the same lines, the complexity of some explanation types can 
hinder users to understand the explanations 
%(e.g., if they lack skills in AI), 
and thus the adoption of the whole ML system (see e.g.~\citet{kaur2020interpreting} in the case of SHAP and GAM). It has also been demonstrated that users assign artificial agents human-like traits, and hence expect these agents to provide explanations using the same conceptual framework they are used to~\cite{graffmalle2017}. Yet,  XAI solutions have not reached such a complete human-centered approach~\cite{miller2019explanation,matarese2021}.

\paragraph{Solutions.}
Mitigating the mismatched explanations somehow obviously relies on applying user-centered approaches when designing XAI systems: two examples include adjusting the types of explanations according to users’ expectations to improve their satisfaction and acceptance \cite{riveiro2022,pazzani2022expert} or integrating a lexical alignment step to improve the understanding of explanations provided by a conversational agent \cite{srivastava2023}. 
Concept-bottleneck models (see e.g. \citet{cbm2023survey,zarlengatabcbm}) may be seen as examples of this family of approaches, when checking that the used concepts are meaningful for the users. 
Another type of solution aims at integrating social and cognitive science’s theories in a theoretical framework for XAI system, e.g. to provide personalized and contextualized explanations~\cite{matarese2021,byrne2023good}. 

%\subsubsection{Counterintuitive explanations}
\subsection{Counterintuitive Explanations}
\label{sec:typology-counterintuitive}
%We then propose to identify counterintuitive explanations when the provided information does not support what users have learned or experienced in the past. 
We then propose to identify \emph{counterintuitive} explanations when the explanation provided by the ML system is in contradiction with the prior knowledge of expert users (i.e., AI practitioners and domain experts). 
Contradiction here occurs at a content level, as opposed to the format level discussed in the case of counterintuitive failures in the previous section.
%As compared to the mismatch failure for which explanations can be contradictory with users expectations in terms of format, counterintuitive failures can be observed when the content of the explanation itself is contradictory with users' knowledge.

\paragraph{Why does it happen?}
Prior knowledge may take various definitions~\cite{dochy1995},  we view it here as  "stored knowledge about the world that have been acquired by an individual"~\cite{brod2013}, including domain expertise and past experience. %Researchers in neuroscience have identified patterns of brain activity that underline the users’ ability to interpret based on past experiences \cite{sohn2019}. 
In an XAI context, %the explanations can differ from such users’ acquired knowledge. 
 users may find faithful explanations %an accurate predictive model's output and a faithful explanations 
contradicting with such prior knowledge, leading them to perceive these explanations as different from what they anticipated ("This is not what I expected" in Fig.~\ref{fig:user-specific}). This is all the more likely to happen as explanations are especially requested when the model's output is perceived by users as abnormal or absurd \cite{riveiro2022}. In the same vein, past experience has been shown to lead to disagreement with the explanations \cite{sohn2019,Suffian2022}. 

\paragraph{Why is it a problem?}
Counterintuitive explanations do not necessarily represent a failure: explanations that do not match user expectations can indeed be used to fix potential issues within the ML system~\cite{kaur2020interpreting} 
or for knowledge discovery~\cite{jimenez2020drug}.
In other situations however, perceiving the provided explanations as counterintuitive can lead users to question the reliability of the prediction even when it is accurate, see e.g.~\citet{Collaris2018,palaniyappan2022} for studies in applied contexts. %the context of insurance fraud and~\cite{} in the context of environmental studies), 
The ML system as a whole may be perceived more negatively~\cite{cabitza2024explanations,nourani2021anchoring,ebermann2023explainable}, potentially impacting users' willingness to engage with AI~\cite{ebermann2023explainable}.
%They can even perceive the explanations as being absurd if the latter do not correspond to their past experiences \cite{sohn2019}.  
This aligns with findings from works in social sciences, 
which have shown that people tend to ignore  information inconsistent with their beliefs from past experiences~\cite{thagard1989,nickerson1998}.%\cite{thagard1989,nickerson1998}% Thus, we believe that the users may perceive some explanations as being absurd as they do not correspond to what they would have expected based on past experiences. 

\paragraph{Solutions.}
Mitigating these failures requires to better align ML explanations and users' prior knowledge. Some works in XAI thus argue for more personalized explanations~\cite{ebermann2023explainable,conati2021toward}  that would directly integrate in their generation the user knowledge, e.g. expressed as features  importance scores~\cite{Jeyasothy2022} or diagrams describing the reasoning process~\cite{lim2025diagrammatization}.
Conversely, the predictive model itself can be changed so the generation of explanations that are aligned with users' prior knowledge is facilitated~\cite{rieger2020,koh2020concept,ross2017right}. In a different perspective, other works propose to co-design explanation interfaces together with experts %in co-design workshops 
so as to integrate both their needs and knowledge \cite{Wang2019,weitz2024explaining}.

%\subsubsection{Biased inferences}
\subsection{Biased Inferences}
~\label{sec:typology-biasedinferences}
Finally, we propose to identify explanation failures relying on \emph{biased inferences} when users make inaccurate interpretations of the explanations, due to cognitive biases.
%(see Fig.~\ref{fig:user-specific}). 
As compared to counterintuitive failures that occur for expert users, for biased inferences we consider mainly lay users who do not have expert knowledge nor past experiences. Yet, such failures can occur for any type of users as cognitive biases are inherent to all humans. 

\paragraph{Why does it happen?}
%The lay users build a mental model of a ML system based on the interaction they have with it~\cite{Liao2020}. Yet, these mental models may be inaccurate as some cognitive biases may interfere in the process \cite{miller2019explanation,bertrand2022,nourani2021anchoring,VANDERWAA2021}.
Similarly to prior beliefs, biases can influence how users respond to different styles of explanations \cite{Liao2020,miller2019explanation}, 
 and several cognitive biases have been shown to trigger inaccurate interpretations of explanations  \cite{bertrand2022,nourani2021anchoring,Wang2019}. We discuss below some common cognitive biases with their consequences on the interpretation of explanations. %\cite{bertrand2022,nourani2021anchoring,VANDERWAA2021}.

\paragraph{Why is it a problem?}
%Biased inferences can lead to misinterpretation of the explanations, which eventually can impair the users' decision-making process \cite{Cheng2019,zhou2021}. %\cite{Cheng2019,cai2019,zhou2021}.
%Several cognitive biases have been shown to trigger inaccurate interpretations of explanations~\cite{bertrand2022}.  %\cite{rudin2018,devisser2020,VANDERWAA2021} %Some cognitive biased can lead to unwanted consequences such as over reliance, under reliance, or misapplication of the explanation \cite{bertrand2022}. 
First, some biases can trigger over reliance in ML explanations. It has been shown that having an explanation, regardless of its quality, increases trust \cite{hoff2015trust,eiband2019impact,lai2019human}.
In other examples, it is shown that longer, richer, explanations are found to be more plausible than shorter ones~\cite{furnkranz2020cognitive,he2025conversational}, and that users may believe they understand better than what they actually do~\cite{rozenblit2002,mueller2019,Chromik2021,he2023knowing}.
% Moreover, the format of these explanations can trigger over reliance: for instance, longer explanations are found to be more plausible than shorter ones \cite{furnkranz2020cognitive}. %; another example is that lay users tend to overrate the depth of their knowledge in XAI \cite{Chromik2021}. %\cite{mueller2019,Chromik2021},
%and can therefore fall into the "illusion of explanatory depth"%: %when interacting with ML explanations: forming false or incomplete interpretations of the explanations and believe they understand better than what they actually do \cite{rozenblit2002,Chromik2021}.  
On the other hand, other biases can trigger under reliance. For instance, the "negativity bias" can cause lay users, in particular, to pay more attention or overweight negative information over positive one of the same strength~\cite{kliegr2021review}.
It may lead users to pay more attention to negative outcomes of the ML system, thus eroding their trust \cite{pratto1991automatic}. %
It has been demonstrated that showing the weaknesses of the system (e.g., competing explanations) or negative outcomes (e.g., a malignant diagnosis) early on can have a major influence on trust \cite{nourani2021anchoring}. 
%Last, 
%it has been observed in various studies that the explanations can be misapplied. Despite the quality of a ML system, many 
Finally, some other biases can trigger users to misapply the explanations: e.g., the "insensitivity to sample size" bias may lead lay users to ignore the statistical significance of a statement \cite{furnkranz2020cognitive}; the "availability" may lead lay users to believe that examples and events that easily come to mind are more representative than is actually the case \cite{Wang2019};
e.g., the "primacy effect" bias may lead them to form an opinion based solely on the first piece of information received \cite{nourani2021anchoring}.%, to cite few examples. 

\paragraph{Solutions.}
Before mitigating these biased inferences, identifying them and measuring their effects on the users' interpretation is a challenging task that can e.g. rely on comparing  users' objective and self-reported understanding~\cite{Cheng2019,bove2022contextualization}. 
Most approaches then rely on the design of appropriate XUI \cite{Wang2019}, for instance controlling what types of predictions users first see when interacting with the system, to mitigate the negative bias \cite{nourani2021anchoring}.

\section{Discussion}
\label{sec:discussion}
The typology of explanation failures presented in the previous section allows to understand why failures happen, how to mitigate them, and how to distinguish them from one another. In this section, we leverage this typology to discuss some key issues, and identify promising research directions.

\subsection{Towards a Holistic XAI Approach}% of the explanation process}
\paragraph{Observation: Some failures result from the interaction between components of the explanation process.}
\label{sec:discussion-1}
Many of the explanation failures discussed in the previous sections
can be diagnosed as stemming from one of these components (model, explainer and user): 
%In the previous section, we delved into the examination of various failures that can occur within an AI system composed of three components: the model, the explainer and the user. The majority of explanation-related issues can often be diagnosed as stemming from one of these components: a technical problem with the system, or a problem with the user's perception and expectations.
they can for instance be due to a technical problem with the system. 
However, our analysis in the previous section 
%reveals another view on the origins of failures: 
also underlines that some errors actually arise from the interplay between the different components considered, rendering them incompatible: %This prompts a new representation of the typology of AI Failures taking into account these interactions, that we propose in Figure~\ref{fig:diagram-interactions}. 
we discussed the interaction between the \textbf{explainer and the user} in Section~\ref{sec:typology_user} and discussed in Section~\ref{sec:typology_syst} some cases of problematic 
%gives rise to problems stemming from the user's prior knowledge and perception of the explanation, such as mismatches and counter-intuitiveness. Furthermore, the previous section highlighted how some system-based failures could be attributed to the user having false expectations on the explanation assumptions (perception of unstability, or competing information), or a lack of knowledge on the assumptions governing the explanation generation.
% On the other side, the 
interaction between the \textbf{model and the explainer}. For instance, %as discussed in Sec.~\ref{sec:typology-unstable}, 
issues like instability may highlight a mismatch between the decision model's behavior (volatile, local), the implicit assumptions made during explanation design (stable boundaries), or the similarity between instances perceived by the user. 

\paragraph{Consequence: subpar technical solutions.}
In the case of these interactions between components, the failure is not caused by a deficiency in either component, but rather from their misalignment.
Consequently, some technical solutions %to address these failures 
suggest adapting or replacing one of the components afterwards, sometimes taking a paradoxical turn: instead of questioning the explainer when observing a failure, it is sometime envisaged instead to train a new AI decision model that would be more adapted to this explainer. As an illustrative example, it has been proposed to build models with more stable behavior to mitigate instability issues of activation-based explainer~\cite{alvarez2018towards}, or models constrained to minimize the disagreement between LIME and SHAP~\cite{schwarzschild2023reckoning}.
This may seem surprising, as explainers are usually leveraged to generate insights about the model, not the other way round.

\paragraph{Going forward: towards a holistic design of the explanation process.}
This problem underscores a significant challenge: the interplay between the elements of the system should be taken into account from the system's inception. The three components of the system should be regarded as interconnected, rather than designed independently. This does not mean abandoning post-hoc methods but rather anticipating their integration in the overall explanation process. 
%This is further confirmed by the paradox behind some proposed technical solutions, that propose, instead of questioning the explainer when observing a failure, training new AI decision models that would be more adapted to the explainer. E.g., models with more stable behavior to mitigate instability issues of gradient-based explainer~\cite{alvarez2018towards}, or models constrained to minimize the disagreement between LIME and SHAP~\cite{azezae}.
%This may seem counterintuitive, as explainers are usually leveraged to generate insights about the model, not the other way around.
Multiple calls for a user-centric approach of XAI have been made, advocating for integrating user needs from the inception of the system and guiding the design of XAI methods~\cite{Wang2019,ribera2019can,vellido2020importance,schmude2023impact}. Nevertheless, more efforts should be pursued on the interaction between AI models and explainers. %, on the other hand, remain limited.
One possibility is to draw inspiration from research in Integrative Design for software systems~\cite{tumer2010integrated}, proposing holistic design strategies for software systems. Design and monitoring of AI systems should thus be conceived in a holistic way, with any choice or change in the decision model prompting a reassessment of its compatibility with the explainer, and vice versa; and similarly for changes in user requirements.

%\subsection{Towards less one-size-fits-all and more transparent XAI solutions}
\subsection{Towards More Transparency and Personalization in XAI Systems}
\label{sec:discussion-2}

\paragraph{Observation: some failures happen because explainers are black-boxes, and both ML practitioners and users ignore their limitations.}
The previous argument about the need to adopt a holistic approach of AI systems design also raises the question of the relevance of one-size-fits-all XAI solutions.
The most well-known XAI methods (e.g. SHAP and LIME) are often conceived under assumptions of data-, model-, and user-agnosticity (i.e. absence of user focus), generally with the aim of providing more flexibility in their use. 
Yet, some of the system-specific failures discussed (see  Sect.~\ref{sec:typology_syst}) and the previous argument about  failures resulting from the interaction of several components suggest that these methods may not, in fact, be one-size-fits-all solutions. 
This adds up to the existing research questioning this user-agnosticity, suggesting that some types of explanations may %in fact 
not be suited to all user profiles~\cite{hoff2015trust,Wang2019}, or to all decision models~\cite{alvarez2018towards,molnar2020general}
%, therefore advocating for self-explaining models).
%Thus, the absolute agnosticity paradigm often considered by XAI researchers should be challenged, for instance by envisaging self-explaining models~\cite{alvarez2018towards} rather than post-hoc explainers. 
This restates that explainers face various limitations and various assumptions, that should be known and understood for proper use.

\paragraph{Consequence: ML practitioners and users misuse XAI systems.}
Unfortunately, these limitations and assumptions of XAI systems are rarely known to users and even ML developers, as analyzed in the proposed typology.
%Besides improper use of XAI methods (e.g. instability issue caused by an impossibility to linearly approximate a black-box model), several failures originate from the lack of knowledge from the side of the user about the assumptions governing the explanation.
While this may seem intuitive for user-specific failures (e.g. overtrust issues in \emph{biased inferences}), it also holds for system-based failures:
for instance, some \emph{competing} issues are caused by the user not knowing how the explainer handles correlations and interactions between features; some \emph{instability} issues by their not understanding the tradeoff between stability and locality, nor what level of locality they wish; some \emph{incompatibility} issues by their not understanding the differences actually captured by two explainers.
Consequently, users may reject or misuse the explanation, not because it does not meet their needs, but because they do not understand how to use it, as shown by ~\citet{kaur2020interpreting}.% for SHAP.

% Observation; des failures viennent du fait que les approches ont des limites qui ne sont pas connues des utilisateurs

% \paragraph{Consequence: ML practitioners and users misuse XAI systems.}
% - conséquence 1: les approhches sont mal utilisées. gros problemes, pas facilement résolubles: car c'est pas juste un bug
% - conséquence 2: les gens vont utiliser des one size fit all vu qu'ils ne sont de toute facon pas conscients des limites possibles. Et ca, ca crée d'autres problemes;

% going forward:
% 1) 
% 2) on est plus transparents dans le design des approches
% 3) on fait plus de XAI literacy
% 4) 

%%%%%%

\paragraph{Going forward: more "transparency" in XAI methods.}
Technical design choices of XAI systems and the assumptions they rely on heavily impact how the explanation should be interpreted.  
Overall, this further confirms the need for more effort in communicating on the core capabilities of XAI systems, e.g. through design principles such as ML transparency \cite{bove2022contextualization}, and more generally improving algorithm literacy of non-ML users~\cite{cabitza2017unintended,chiang2022exploring,he2025conversational}. 
%, as these failures result from improper use of an explanation method by the ML practitioner or the user being unaware of some
%In the case of counterfactual explanations, \cite{laugel2023achieving} argue that besides being transparent on the criteria optimized, explanation methods should be transparent on how they combine them, as they are often at odds with one another, despite the user not knowing.
%Efforts 
This also aligns with prior calls to further encourage interdisciplinary works for designing more transparent XAI systems, instead of building  black-box explainers.
%However, research in this remains timid and more efforts should be made. 
One possible direction to pursue is improving the standardization of XAI methods~\cite{haque2023explainable}, e.g. through the description of explanation methods in a factual way, akin to Model Cards~\cite{mitchell2019model}.

%to circumvent some failures such as unstability, the technical solutions proposed training new AI decision models that would be more adapted to the explainer
%Yet, as some failures result from an inadequacy of such explainers to some models (cf Fig~\ref{fig:diagram-interactions}), the relevance of such one-size-fits-all XAI solutions is questioned.
%Some of the existing technical solutions propose training new AI decision models that would be more adapted to the explainer. E.g., models with more stable behavior to mitigate instability issues of gradient-based explainer~\cite{alvarez2018towards}, or models constrained to minimize the disagreement between LIME and SHAP~\cite{azezae}.
%As a result, instead of questioning the explainer, some of the existing technical solutions propose training new AI decision models that would be more adapted to the explainer. E.g., models with more stable behavior to mitigate instability issues of gradient-based explainer~\cite{alvarez2018towards}, or models constrained to minimize the disagreement between LIME and SHAP~\cite{azezae}. This may seem counterintuitive, considering the explainer are used to generate insights about the model, not the other way around.

\subsection{Towards More XUI}
%\subsection{User interfaces can help mitigate the harms of explanation failures}
\label{sec:discussion-3}
\paragraph{Observation: some failures happen because users are limited in their interaction with explanations.}
From a cognitive point of view, the explanation process is argued to be interactive~\cite{miller2019explanation} and yet, at best, it is merely sequential in XAI. Most approaches are limited to the generation of factual information about the model's behaviour or the predicted outcome (e.g., feature importance scores, counterfactual examples, etc.) and users are often not able to interact with it.
%V1
%As discussed in the previous sections, most XAI methods present the users with rather factual information about the model's behaviour or the predicted outcome (e.g., feature importance scores, counterfactual examples, etc.). Explaining ML models remains a complex task and the insights provided by such explanations is limited by the capacities of XAI methods to generate faithful information. Hence, the latter can be limited: it concerns mostly the predictive model and does not take into account the users' needs or knowledge; it does not allow the users to interact with it.
%V2
%Explaining ML models remains a complex task and the insights provided by such explanations is limited by the capacities of XAI methods to generate faithful information about the predictive model: these methods do not take into account the users' needs or knowledge; it does not allow the users to interact with it.

\paragraph{Consequence: external information and processes interfere with the explanations. }
%causing 
This causes 
the explanations to be potentially perceived by users as  incomplete or incorrect (e.g., lack of knowledge or lack of transparency on the XAI method). As discussed in Sect.~\ref{sec:typology-biasedinferences}, users may thus draw on external information and cognitive processes to interpret the explanations, potentially leading to  
%one or more 
XAI failures. For instance, users may expect that there can be intuitive changes in counterfactual examples because they have experienced the same logical path in real life \cite{Suffian2022}.
%This causes the explanations to be inconsistent from a ML point of view (e.g., lack of stability or competing explanations) and incomplete from a user point of view (e.g., lack of knowledge or lack of transparency on the XAI method). As a consequence, one or more explanation failures are more likely to occur. These limitations may result in having the users using other information and different complex cognitive processes to interpret the explanations. As discussed in the previous Section, some users may refer to some external knowledge or beliefs: %For instance, users with specific beliefs (e.g., the Earth is flat) can be influenced by the confirmation bias when analyzing information (e.g., only looking for information that confirms that the Earth is flat). 

\paragraph{Going forward: more user interfaces for explanations.}
%\paragraph{Going forward: a personalized approach to explaining ML systems}
%From a cognitive point of view, \cite{miller2019explanation} emphasizes the importance of the social dimension in the explanation process. 
%It has been discussed in the previous section that adding users' knowledge onto the explanations could help mitigate counterintuitive failures.
%In a "human-in-the-loop" paradigm, it would be interesting to go deeper into personalization. By collecting users' needs in an automated way, the explanations can be tailored to each of them. 
%The explanation can be designed as a dialog with real interactions between the two parts \cite{miller2019explanation}.
%To mitigate these harms, a research perspective could consider a more personalized approach for explaining ML systems. It has been discussed in the previous section that adding users' knowledge onto explanations could help mitigate counterintuitive failures. For example, when the objective of an explanation is to optimize the prediction, a counterfactual example may propose a change that is usually feasible, but a user may not be able to do it because of his/her personal situation  \cite{poyiadzi2021overlooked}. Personalizing the explanations would thus allow to offer users with only the information they need. The question of how to collect these needs arises. 
%We believe that investigating new interactive and dynamic design enhancements to collect such needs in XUIs would contribute to provide personalized and useful ML explanations. 
We believe that XUIs allow to better organize, complete and display ML explanations according to the user needs~\cite{chromik2021human}. Moreover, accounting for the users is key to design such interfaces, which thus forces AI practitioners to adopt a user-centered approach when conceiving ML systems (see Sect.~\ref{sec:discussion-1}). Previous studies have demonstrated the usefulness of visual interfaces to present ML explanations (see e.g., \citet{szymanski2021visual,ooge2022explaining})
but there are other interaction modalities, in particular the conversational mode. As users have progressively become more familiar with conversational AIs (e.g. ChatGPT), the provided explanations may  be presented in conversational interfaces and become a  dialog, i.e. questions from the users and corresponding answers from the machine \cite{miller2019explanation,ribera2019can}. This could open new perspectives for design principles such as introducing a narrative logic that allows temporizing the current information. For instance primary information can be controlled to mitigate the negativity bias~\cite{nourani2021anchoring}. 
We believe that studying such a modality for the display of explanations would allow to better understand users' processes for analyzing and understanding ML explanations. 

\section{Conclusion}
%The typology of explanation failures presented in the previous section allows to understand why failures happen, how to mitigate them, and how to distinguish them from one another.
In this work, we have proposed a typology of XAI failures allowing to understand why failures happen, how to mitigate them, and how to distinguish them from one another. We believe it can help AI developers and designers better understand XAI systems and their limitations. Leveraging this typology, we have identified promising research avenues for XAI. In addition to these directions, future works will include investigating the potential interaction between multiple failures. Besides better understanding their connections to one another, the co-occurrence or superposition of several failures raises crucial questions regarding their consequences on user understanding. Furthermore, a better understanding of each failure and these interactions could be leveraged to formally propose a diagnostic framework to help identifying their origins in the explanation process.

\bibliography{biblio}

% \newpage

% \appendix

% \section{Details on the Methodology}

% \subsection{Ending conditions}
% As explained in Section~\ref{sec:method}, the process of building the typology stops when two conditions are met: first, the definition of the dimensions and characteristics (types of XAI failures) needs to be stable over the last iteration of papers. Then, the typology needs to be deemed concise, robust, comprehensive and explanatory. We define these terms the following way:

% \begin{itemize}
%     \item concise: the typology contains 7 types of failures, which is deemed concise enough
%     \item robust: the typolog
% \end{itemize}

% \subsection{Lists of keywords considered}

% This section details the keywords that were consi

\end{document}